\documentclass{article}

\usepackage{microtype}
\usepackage{graphicx}
\usepackage{subfigure}
\usepackage{booktabs}
\usepackage[accepted]{icml2025}
\usepackage{xurl}

\icmltitlerunning{Does Welsh Media Need a Review? Detecting Bias in Nation.Cymru's Political Reporting}

\begin{document}

\twocolumn[
\icmltitle{Does Welsh Media Need a Review?\\
Detecting Bias in Nation.Cymru's Political Reporting}

\icmlsetsymbol{equal}{*}

\begin{icmlauthorlist}
\icmlauthor{Cai Parry-Jones}{cam}
\end{icmlauthorlist}

\icmlaffiliation{cam}{University of Cambridge, Cambridge, United Kingdom}

\icmlcorrespondingauthor{Cai Parry-Jones}{ctp27@cam.ac.uk}

\icmlkeywords{media bias, NLP, sentiment analysis, LLM, Welsh politics}

\vskip 0.3in
]

\printAffiliationsAndNotice{}
\setcounter{footnote}{1} 

\begin{abstract}
Wales' political landscape has been marked by growing accusations of bias in Welsh media. This paper takes the first computational step toward testing those claims by examining Nation.Cymru, a prominent Welsh political news outlet. I use a two-stage natural language processing (NLP) pipeline: (1) a robustly optimized BERT approach (RoBERTa) bias detector for efficient bias discovery and (2) a large language model (LLM) for target-attributed sentiment classification of bias labels from (1). A primary analysis of 15,583 party mentions across 2022–2026 news articles finds that Reform UK attracts biased framing at twice the rate of Plaid Cymru and over three times as negative in mean sentiment ($p<0.001$). A secondary analysis across four parties across both news and opinion articles shows that Plaid Cymru is the outlier, receiving markedly more favourable framing than any other party. These findings provide evidence of measurable differential framing in a single Welsh political media outlet, supporting calls for a broader review of Welsh media coverage. Furthermore, the two-stage pipeline offers a low-cost, replicable framework for extending this analysis to other Welsh outlets, as well as media ecosystems outside of Wales.
\end{abstract}

\section{Introduction}

The run-up to the 2026 Welsh Parliament (Senedd) elections has been marked by growing accusations of political bias in Welsh media. Representatives from both the Conservatives and Reform UK have publicly questioned the impartiality of political news coverage in Wales.\footnote{BBC News, ``Reform calls for review into BBC and Plaid relationship'', 11 November 2025.}\textsuperscript{,}\footnote{Nation.Cymru, ``Tory MS raises concerns over `impartial' media coverage of Senedd election'', 21 February 2026.} But are these complaints supported by evidence?

This paper offers the first computational approach to answering this question by examining Nation.Cymru, a prominent and award-winning\footnote{Journalists' Charity, Wales Media Awards 2025, 17 January 2026.} Welsh political news outlet. Nation.Cymru describes itself as an independent, national news service that reports on ``all political parties without fear or favour.''\footnote{Nation.Cymru, ``Our only bias is in favour of truth and honesty and against those who lie and deceive'', 6 December 2025.}

With polling placing Reform UK and Plaid Cymru as the two leading parties, I focus on whether Nation.Cymru's coverage exhibits measurable differential sentiment bias between them and other parties. If bias is detectable in this outlet, it strengthens the case for a broader review of Welsh political media. Furthermore, with the next Senedd controlling a \pounds27bn budget, media bias is not just an electoral concern, it is an economic one \citep{besley2006, larcinese2011}.

\section{Literature Review}

\citet{dalessio2000} distinguish three forms of media bias: coverage bias (how much attention an outlet gives each party), gatekeeping bias (whether stories are selectively covered or ignored), and statement bias (whether that coverage is favourable or unfavourable). Coverage bias and gatekeeping bias are difficult to measure with a single outlet, and recent research suggests that statement bias is the most consequential dimension for electoral outcomes \citep{eberl2017}. This paper therefore focuses on statement bias.

Systematic reviews confirm that NLP methods can increasingly detect media bias \citep{rodrigogines2024, spinde2023, kuila2024}. For sentence-level bias detection, \citet{spinde2021} introduced the BABE expert-annotated dataset, on which \citet{ghosh2025} fine-tuned the \texttt{himel7/bias-detector} RoBERTa model used in this project. LLMs can match or exceed fine-tuned classifiers on sentiment tasks, but at substantially higher computational cost \citep{shao2025}. Model cascading, where a lightweight model handles simple cases and a costlier model tackles harder ones, can reduce costs without a noticeable reduction in accuracy \citep{yue2024}.

\section{Methodology}

\paragraph{Stage~1: Bias detection.}
Each mention's context window is classified as \textsc{biased}/\textsc{non-biased} using \texttt{himel7/bias-detector}. This stage is lightweight enough to classify all mentions locally, passing only those labelled containing loaded language forward to Stage~2.

\paragraph{Stage~2: Target attribution and sentiment analysis.}
A key challenge is distinguishing bias \textit{about} a party from sentiment \textit{expressed by} a party (e.g.,\ ``Plaid criticised the devastating policy'' has negative language, but Plaid is the speaker, not the target). I initially tested a dedicated sentiment model but found it lacked the nuanced reasoning needed to distinguish speaker from target.\footnote{\texttt{cardiffnlp/twitter-roberta-base-sentiment-\allowbreak latest}, a RoBERTa model fine-tuned on ${\sim}124$M tweets for three-class sentiment. See \url{https://huggingface.co/cardiffnlp/twitter-roberta-base-sentiment-latest}.} I therefore use the LLM Claude Sonnet 4.6 \citep{anthropic2026} via API as a prompted classifier (prompt available in the code repository). For each biased mention it returns: (1)~\texttt{on\_target} (1/0): is the bias directed at the named party (off-target mentions are excluded); (2)~sentiment score on a 5-point categorical scale $s \in \{-1, -0.5, 0, +0.5, +1\}$; (3)~one-sentence reasoning. The prompt includes speaker-vs-target examples and a self-check instruction. To validate Stage~2 output, I manually reviewed a stratified random sample of 50 classified mentions and agreed with the LLM's target-attribution and sentiment labels in 92\% of cases; the LLM occasionally incorrectly labeled the on-target party when multiple parties appeared in the same context window.

\paragraph{Design choices.} Context window $\pm 1$ was selected after testing windows of 0, 1, and 2 sentences around each mention (see code repository); larger windows increased the bias detection rate, consistent with findings that transformer classifiers are sensitive to surrounding context for bias identification \citep{spinde2021}. However, larger context windows lead to higher costs for Stage~2 processing; $\pm 1$ was deemed a good middle ground. The LLM prompt uses a reasoning-first output format (reasoning before classification fields), which improves classification quality. The dataset is large enough that incomplete bias detection does not preclude statistical significance.

\paragraph{Implementation.} Stage~1: run locally on an M3 MacBook Air. Stage~2: Anthropic API; total cost was approximately \$6 for the primary analysis and \$5 for the secondary. For the secondary analysis, stage 2 was capped at 250 randomly selected biased mentions per party per article type. Without this cap, the secondary analysis would have cost an estimated \$40+. All code and data are available at \url{https://github.com/caitpj/d200_media_bias}.

\begin{table*}[t]
\caption{Primary analysis: Reform UK vs Plaid Cymru, news articles, 2022-2026.}
\label{tab:primary}
\centering\small
\begin{tabular}{lcc}
\toprule
 & \textbf{Reform UK} & \textbf{Plaid Cymru} \\
\midrule
Total mentions & 7{,}237 & 8{,}346 \\
On-target biased (\% of total) & 1{,}124 (15.5\%) & 629 (7.5\%) \\
Mean sentiment & $-0.735$ & $-0.211$ \\
Strongly negative (\% of total) & 668 (9.2\%) & 147 (1.8\%) \\
\midrule
Bias rate $z$-test & \multicolumn{2}{c}{$z = 15.75$, $p < 0.001$, $h = 0.25$} \\
Sentiment $t$-test & \multicolumn{2}{c}{$t = -20.93$, $p < 0.001$, $d = -1.04$} \\
\bottomrule
\end{tabular}
\end{table*}

\section{Data}

Nation.Cymru does not provide a structured database for research access. I therefore collected articles via its WordPress REST API, parsing the returned HTML into plain text. To avoid irrelevant articles, only those containing at least two Welsh politics keywords (e.g.,\ `Senedd', party names, notable political figures; see code repository for full list) were requested. This yields 11{,}847 articles spanning 2017-2026.

I extract individual party mentions via string matching, with a $\pm 1$ sentence context window around each. Sentences shorter than 30 characters (photo captions, stubs) are excluded to prevent double-counting of context already captured by adjacent mentions. The primary analysis restricts to Reform UK and Plaid Cymru, 2022-2026 (the inter-election period following the 2021 Senedd election), news articles only, yielding 15{,}583 mentions (7{,}237 Reform UK; 8{,}346 Plaid Cymru). The secondary analysis covers four parties: Conservatives, Labour, Reform UK, and Plaid Cymru, 2022-2026 across news and opinion articles (250 randomly selected biased mentions per party per article type from 57{,}938 mentions).

\section{Results}

The primary analysis compares Reform UK and Plaid Cymru in news articles. Reform attracts on-target biased framing at twice Plaid's rate and over three times the negative sentiment, a large and statistically significant gap ($p < 0.001$; Table~\ref{tab:primary}).

\begin{figure}[ht]
\centering
\includegraphics[width=\columnwidth]{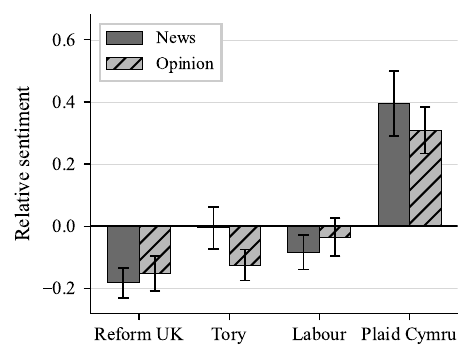}
\caption{Secondary analysis: Mean sentiment relative to article-type mean (news: $-0.590$; opinion: $-0.479$) by political party, 2022-2026. Note, negative baselines are typical of political news coverage \citep{thesen2024}. Error bars show 95\% confidence intervals.}
\label{fig:secondary}
\end{figure}

The secondary analysis extends to four parties across news and opinion articles (Figure~\ref{fig:secondary}). The most striking pattern is that Plaid Cymru is a clear positive outlier, receiving markedly more favourable framing than any other party in both news and opinion articles. Reform UK, the Conservatives, and Labour all cluster below the article-type mean sentiment score. Despite being the only party omitted from Nation.Cymru's founding charter, Reform UK is not framed significantly differently from Labour and the Conservatives.

\section{Discussion}

\paragraph{Findings.} These findings are consistent with the complaints discussed in the Introduction, and provide the first computational evidence of differential framing in Welsh political media. The secondary analysis suggests that the dominant pattern is not that Reform UK is singled out for negative treatment, but that Plaid Cymru receives uniquely favourable framing, a finding at odds with the outlet's claim to report on all parties without favour. However, these findings only reflect a single outlet; whether this pattern extends across Welsh media remains an open question, but the results presented here suggest that a broader review is warranted.
On the methodology side, the LLM target-attribution step is critical for valid statement bias measurement: without it, 33.4\% of biased mentions would be misattributed. This underscores the capacity of contemporary LLMs to handle nuanced classification tasks that exceed the capabilities of fine-tuned specialist models. More broadly, the two-stage design offers a low-cost, replicable framework for measuring media bias in other outlets and media ecosystems.

\paragraph{Limitations.} Coverage bias and gatekeeping bias are dimensions of media bias \citep{dalessio2000}, but have not been assessed (requires data from other outlets). LLM outputs are stochastic, so exact replication of results is not guaranteed, though manual validation suggests high classification reliability. When both parties appear in the same context window, the LLM occasionally assesses the more salient party rather than the one it was asked about (${\sim}6\%$ of on-target classifications). Keyword matching cannot handle coreference, missing anaphoric references to previously mentioned parties (e.g., ``the party's reckless decisions''), meaning some mentions will have been missed from the analysis. Statistical tests treat each mention as independent, but mentions from the same article are likely correlated. This could result in a few highly biased articles making up a disproportionate number of mentions.

\paragraph{Next steps.} Extending to other Welsh outlets (BBC Wales, WalesOnline) would establish whether this pattern is specific to Nation.Cymru or shared across Welsh media, and would also enable measurement of coverage bias and gatekeeping bias for individual outlets. With a larger budget, the 250-mention random sampling cap in the secondary analysis could be removed. Clustering standard errors by article would account for within-article correlation and provide more robust inference. Delving into the full dataset, 2017-, might present other interesting insights such as bias frequency and intensity changes in the run-up to elections. 

\bibliographystyle{icml2025}
\bibliography{references}

\end{document}